\documentclass{article}
\usepackage{spconf,amsmath,graphicx}
\usepackage{booktabs}
\usepackage{pifont}
\usepackage{makecell}
\title{A Dual-Scale Lead-Separated Transformer with Lead- \\ Orthogonal Attention and Meta-Information for ECG Classification}
\name{Yang Li\textsuperscript{1}, Guijin Wang\textsuperscript{1}, Zhourui Xia\textsuperscript{2}, Wenming Yang\textsuperscript{3}, Li Sun\textsuperscript{1}\sthanks{Corresponding author: sunli\_3595@163.com.}} % The work is supported by NSFC from PRC (1234567890) and Hunan NSF (1234567890)}}
\address{\textsuperscript{1}Department of Electronic Engineering, Tsinghua University, Beijing 100084, China, \\ \textsuperscript{2}Huachuang Aima Information Technology, Chengdu 610094, China,  \textsuperscript{3}Department of Electronic \\ Engineering, Tsinghua Shenzhen International Graduate School, Shenzhen 518055, China}

\begin{document}
%\ninept
%
\maketitle
\begin{abstract}  % 100-150 words

Auxiliary diagnosis of cardiac electrophysiological status can be obtained through the analysis of 12-lead electrocardiograms (ECGs). This work proposes a dual-scale lead-separated transformer with lead-orthogonal attention and meta-information (DLTM-ECG) as a novel approach to address this challenge. ECG segments of each lead are interpreted as independent patches, and together with the reduced dimension signal, they form a dual-scale representation. As a method to reduce interference from segments with low correlation, two group attention mechanisms perform both lead-internal and cross-lead attention. Our method allows for the addition of previously discarded meta-information, further improving the utilization of clinical information. Experimental results show that our DLTM-ECG yields significantly better classification scores than other transformer-based models, matching or performing better than state-of-the-art (SOTA) deep learning methods on two benchmark datasets. Our work has the potential for similar multichannel bioelectrical signal processing and physiological multimodal tasks.

\end{abstract}

\section{Introduction}
\label{sec:intro}

\begin{figure*}[ht]
\centering

\begin{minipage}[b]{.48\linewidth}
  \centering
  \centerline{\includegraphics[height=5cm]{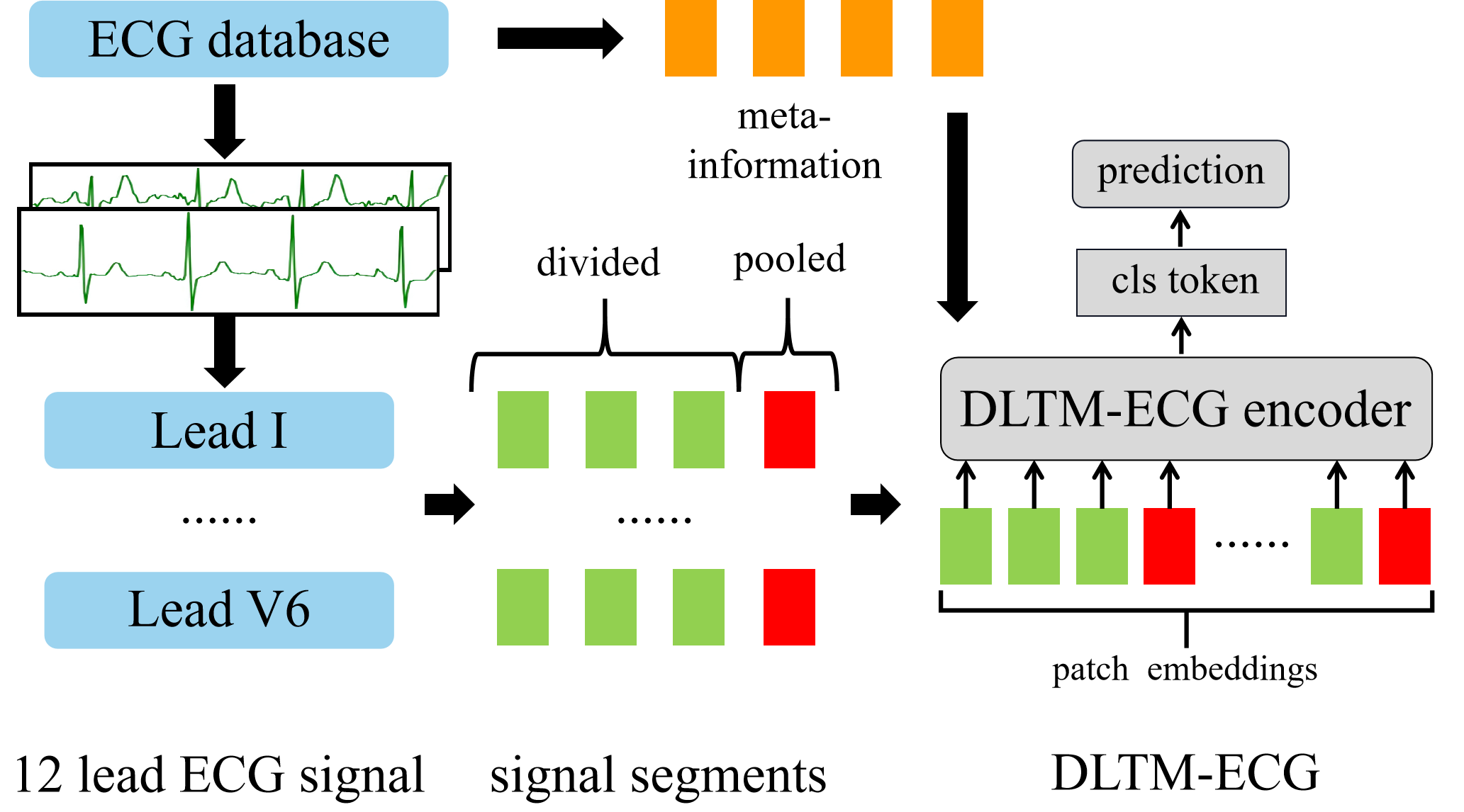}}
%  \vspace{1.5cm}
  \centerline{(a) Input signal is embedded to patches as input. }\medskip
\end{minipage}
\hfill
\begin{minipage}[b]{0.48\linewidth}
  \centering
  \centerline{\includegraphics[height=5.5cm]{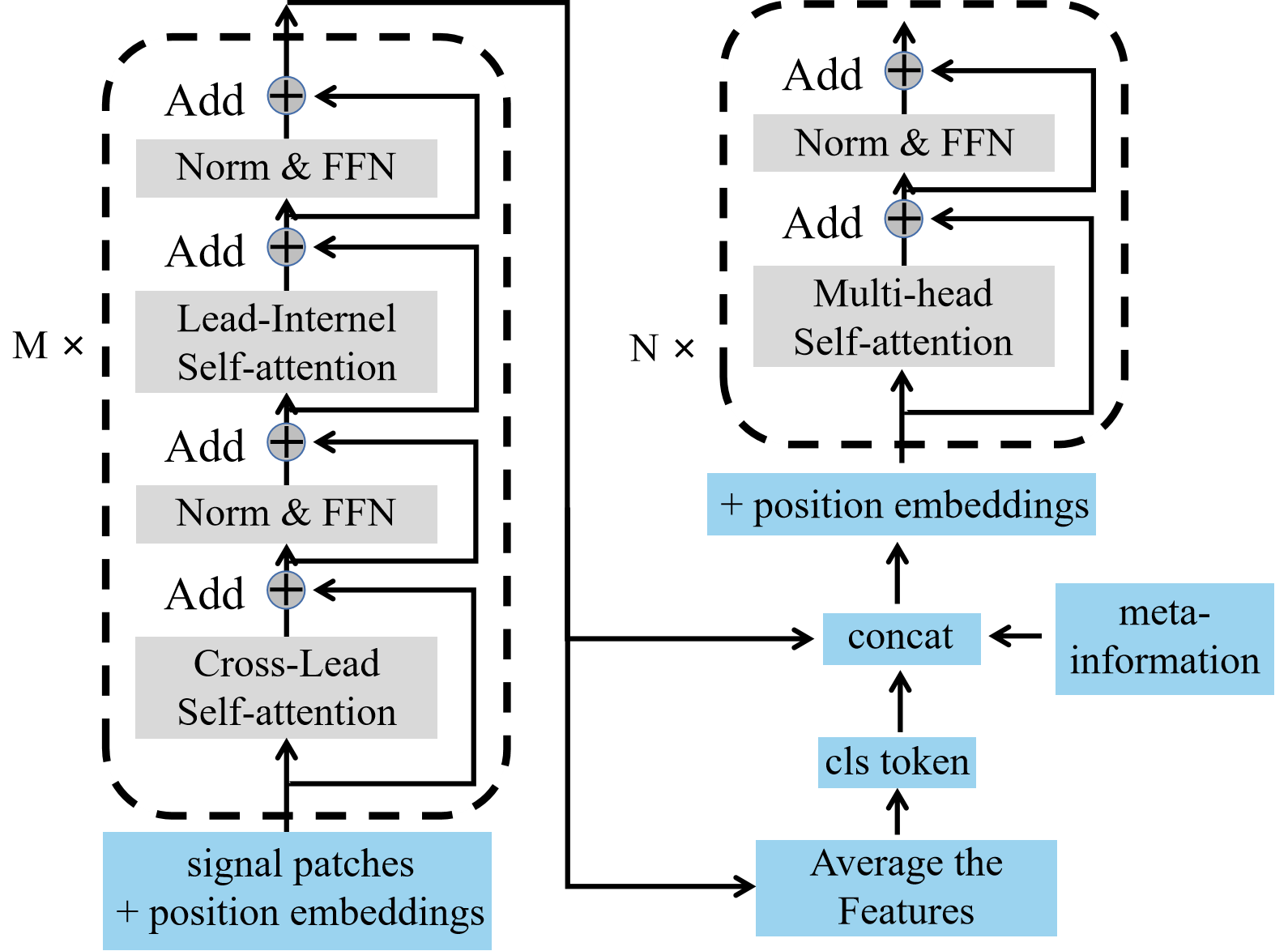}}
%  \vspace{1.5cm}
  \centerline{(b) Encoder structure.}\medskip
\end{minipage}

\caption{Overview of DLTM-ECG method.}
\label{fig:model_structure}
\end{figure*}

% 引用方法：
% \textbf{Figure \ref{fig:res}}

Cardiovascular disease (CVD) remains a leading cause of death worldwide, greatly necessitating improvements in the quality and quantity of care \cite{vos2020global}. Cardiac electrophysiological recordings have become fundamental for discerning cardiac symptomology in patients, and a plethora of information can be examined through the interpretation of 12-lead electrocardiograms (ECGs). More accessible medical services and the widespread use of wearable ECG monitoring devices have greatly increased the availability of ECG data for clinicians, but simultaneously it has become impossible for cardiologists to manually interpret the massive amounts of data generated in these scenarios. 

In recent years, automatic ECG diagnostic technology has been developed based on custom-built deep learning models such as Convolution Neural Networks (CNNs), Long Short-Term Memory networks (LSTMs), and transformer-based networks \cite{reddy2021imle,strodthoff2020deep,li2021bat,mehari2022self}, some of which report current state-of-the-art (SOTA) scores. Currently, researchers involved with automated ECG diagnostics are paying increased attention to transformer-based models which have performed very well in a variety of fields due to transformers’ advantages in capturing long-range attention and ability to encode input with various semantics \cite{vaswani2017attention,nagrani2021attention}. In a recent study, CNN layers and transformer layers extracted information directly from ECG segments, and 20 of 300 features designed by experts were selected including heart rate variability and morphological features \cite{natarajan2020wide}. BaT regards the aligned heartbeat sequence as input and aggregates all heartbeat features with the convolution layer after 1D Swin transformer blocks \cite{li2021bat}.

In this work, we propose a novel lead-separated transformer with meta-information called DLTM-ECG, with dual-scale representation and lead-orthogonal attention mechanisms. In this proposed neural network, the one-dimensional multichannel ECG input is first subdivided into patches by time and lead, then concatenated with reduced-dimension features to obtain dual-scale information. We adopt not only vanilla attention but also lead-orthogonal attention mechanisms. To adaptively integrate information from all feature tokens, the global average pooling (GAP) of features serves as the initialization of the class token. The meta-information from the ECG database is encoded and adopted to facilitate comprehensive learning about the representation of different types of information. Experimental results show that our method is significantly improved over the previous ECG transformer structure and matches or performs better than the previous state-of-the-art (SOTA) methods in multiple tasks. 
% This method can provide a reference for one-dimensional bioelectrical signal classification and physiological multimodal tasks related to ECG.

\section{Methodology}

\subsection{Model structure}

% In designing a transformer method suitable for ECG classification, we considered the following ECG characteristics as the intuitive basis for our model. 
% %Ablation experiments in 4.2 then confirmed the effectiveness of our unique design. 
% First, different ECG leads are regarded by cardiologists as diverse perspectives of cardiac electrical activity [5], and there is an association between multiple channels. Our approach requires a balanced trade-off between modeling the information in each single lead and establishing the correlation among all the leads. 
% Second, during ECG diagnosis, some symptoms need to be judged by coarse-grained features such as RR interval [3,18], while others require attention to more granular local abnormalities. Although the transformer method has advantages in capturing long-range attention, it does have difficulty synthesizing information at different scales. Self-attention performed on a single scale will lead to accuracy degradation. 
% Finally, ECG has a quasi-periodicity with heart rate as the frequency, and where morphology of the different leads is similar. When adapting transformer blocks with vanilla self-attention, some segments, although located in different leads and different time periods, may have similar characteristics, leading the model to assign attention to uninterested regions and introduce interference noise. 

\textbf{Fig. \ref{fig:model_structure}} shows the overall structure of the proposed model. 
We obtained both ECG signals and meta-information from public databases. 
Input ECG is divided by time and lead. 
The signal from each lead is pooled to obtain reduced-dimension features with a coarse scale which, together with the original signal, constitutes a dual-scale representation. 
The initial few layers of the model adopted lead-orthogonal attention modules to define the computational range of attention and reduce interference from segments with low correlation. 
The global information is then integrated in transformer layers with vanilla attention. 
\textbf{Table \ref{table: hyperparameters}} shows some model hyperparameters. DLTM-ECG has 2.6M trainable parameters.
Code will be available.
% Some model structure hyperparameters were set as shown in . 
% To adaptively integrate information from all tokens, the class token is initialized by the global average pooling (GAP) of previous features. 
% Tokens representing meta-information are concatenated to deep features before vanilla attention. 

\begin{table}\centering
\caption{Model hyperparameters. }
  \begin{tabular}{cc}
  
    \toprule[1.5pt]
    %%%%%%%%%%%%%%%%%%%%%%%%%%%%%%%%%%%%%%%%%%%%%
    %\textwidth 是每一行的宽度.[0.1\textwidth]设定单元格宽度
    % [c]  单元格文本居中对齐
    % {name} 单元格内容
    %%%%%%%%%%%%%%%%%%%%%%%%%%%%%%%%%%%%%%%%%%%%%
    \makebox[0.3\textwidth][c]{\makecell[c]{Input size}} &
    \makebox[0.1\textwidth][c]{\makecell[c]{(12, 250)}}  
    \\
    \midrule[1pt]
    \makebox[0.3\textwidth][c]{\makecell[c]{ECG segment length}} &
    \makebox[0.1\textwidth][c]{\makecell[c]{50}}   
    \\
    \midrule[1pt]
    \makebox[0.3\textwidth][c]{\makecell[c]{Embedding dimension}} &
    \makebox[0.1\textwidth][c]{\makecell[c]{160}} 
    \\
    \midrule[1pt]
    \makebox[0.3\textwidth][c]{\makecell[c]{Hidden dimension}} &
    \makebox[0.1\textwidth][c]{\makecell[c]{480}} 
    \\
    \midrule[1pt]
    \makebox[0.3\textwidth][c]{\makecell[c]{Number of lead-orthogonal \\ attention blocks}} &
    \makebox[0.1\textwidth][c]{\makecell[c]{4}}  
    \\
    \midrule[1pt]
    \makebox[0.3\textwidth][c]{\makecell[c]{Number of MSA blocks}} &
    \makebox[0.1\textwidth][c]{\makecell[c]{2}} 
    \\
    \midrule[1pt]
    \makebox[0.3\textwidth][c]{\makecell[c]{Number of heads}} &
    \makebox[0.1\textwidth][c]{\makecell[c]{5}} 
    \\
    % \midrule[1pt]
    % \makebox[0.3\textwidth][c]{\makecell[c]{Trainable parameters}} &
    % \makebox[0.1\textwidth][c]{\makecell[c]{2.6 M}} 
    % \\
    \bottomrule[1.5pt]
  \end{tabular}
  \label{table: hyperparameters}
\end{table}

\subsection{Lead-separated ECG segmentation on dual scales}

% First, different ECG leads are regarded by cardiologists as diverse perspectives of cardiac electrical activity [5], and there is an association between multiple channels. Our approach requires a balanced trade-off between modeling the information in each single lead and establishing the correlation among all the leads. 
% Second, during ECG diagnosis, some symptoms need to be judged by coarse-grained features such as RR interval [3,18], while others require attention to more granular local abnormalities. Although the transformer method has advantages in capturing long-range attention, it does have difficulty synthesizing information at different scales. Self-attention performed on a single scale will lead to accuracy degradation. 
Different ECG leads are regarded by cardiologists as diverse perspectives of cardiac electrical activity \cite{liu2018multiple}.
Our transformer-based model is expected to notice diversity among various ECG leads, hence we proposed a lead-separated division method. Specifically, a segment from one single channel is treated as a patch. At the same time, since it is difficult for transformer blocks to synthesize information at different scales, we pooled ECG signals in the time dimension to introduce both coarse and fine-grained scales. As a commonly used method to reduce feature dimension, maximum pooling could retain peak and elevation information and demonstrate sufficient effectiveness in our experiments. All patches are linearly projected into patch embeddings as inputs of the DLTM-ECG encoder. We used 12-lead ECG as an example in this work, but due to the transformer's flexibility in the number of input patch embeddings, our model can support records with a variable number of leads. 

% When pooling ECGs we were faced with two choices- maximum pooling or average pooling, both of which are commonly used methods to reduce feature dimensions. We ultimately selected maximum pooling as shown in Figure 2(b) because ECG signals contain important high-frequency information. The QRS complex undergoes a dramatic change of positive and negative peaks within 0.1 seconds and furthermore, abnormal elevation in some specific locations such as ST-segment may suggest underlying cardiac disease. These features are better retained by maximum pooling. In contrast, average pooling tends to drastically smooth varying amplitude values and small elevations, resulting in a waveform close to the horizontal line which can lead to loss of important features. 

% The overall concept described in this section is an innovation for ECG classification to excavate abundant, scalable features in different leads. By contrast, ViT for ECG extends the method of image classification by encoding all channels together without considering granularity differences. Swin for ECG aggregates local information in a progressive way. BaT nests beat aggregation blocks to learn global features. In our model, we determined that downsampling is an effective way to simultaneously capture features in both granularities while retaining novelty among ECG classification methods. In such manner, our method retains some feature similarity and facilitates attention calculation.

\subsection{Multiple attention calculation mechanisms}

\begin{figure}[ht]
\centering

\begin{minipage}[b]{.48\linewidth}
  \centering
  \centerline{\includegraphics[height=6.0cm]{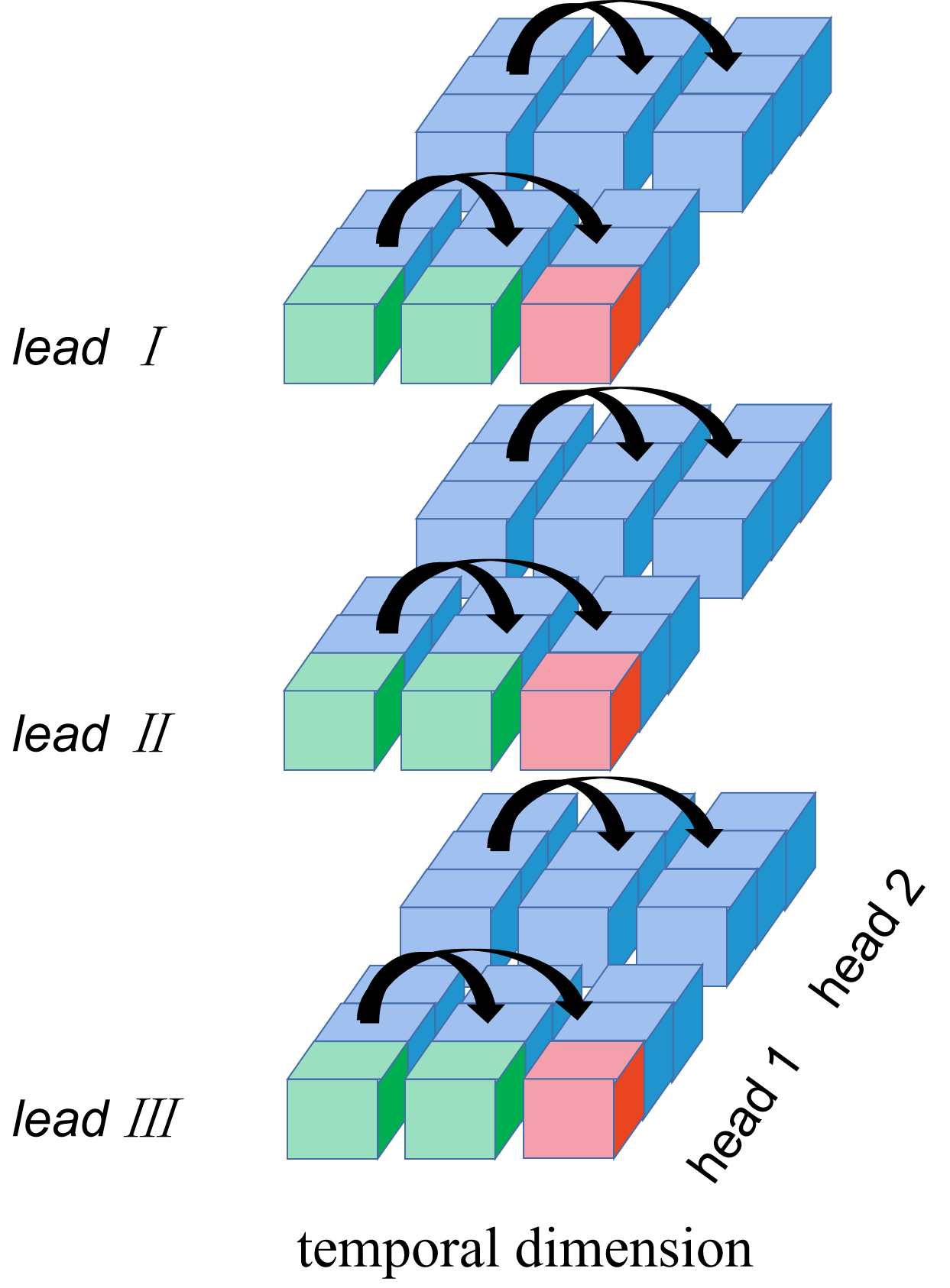}}
%  \vspace{1.5cm}
  \centerline{(a) Lead-internal self-attention.}\medskip
\end{minipage}
\hfill
\begin{minipage}[b]{0.48\linewidth}
  \centering
  \centerline{\includegraphics[height=5.9cm]{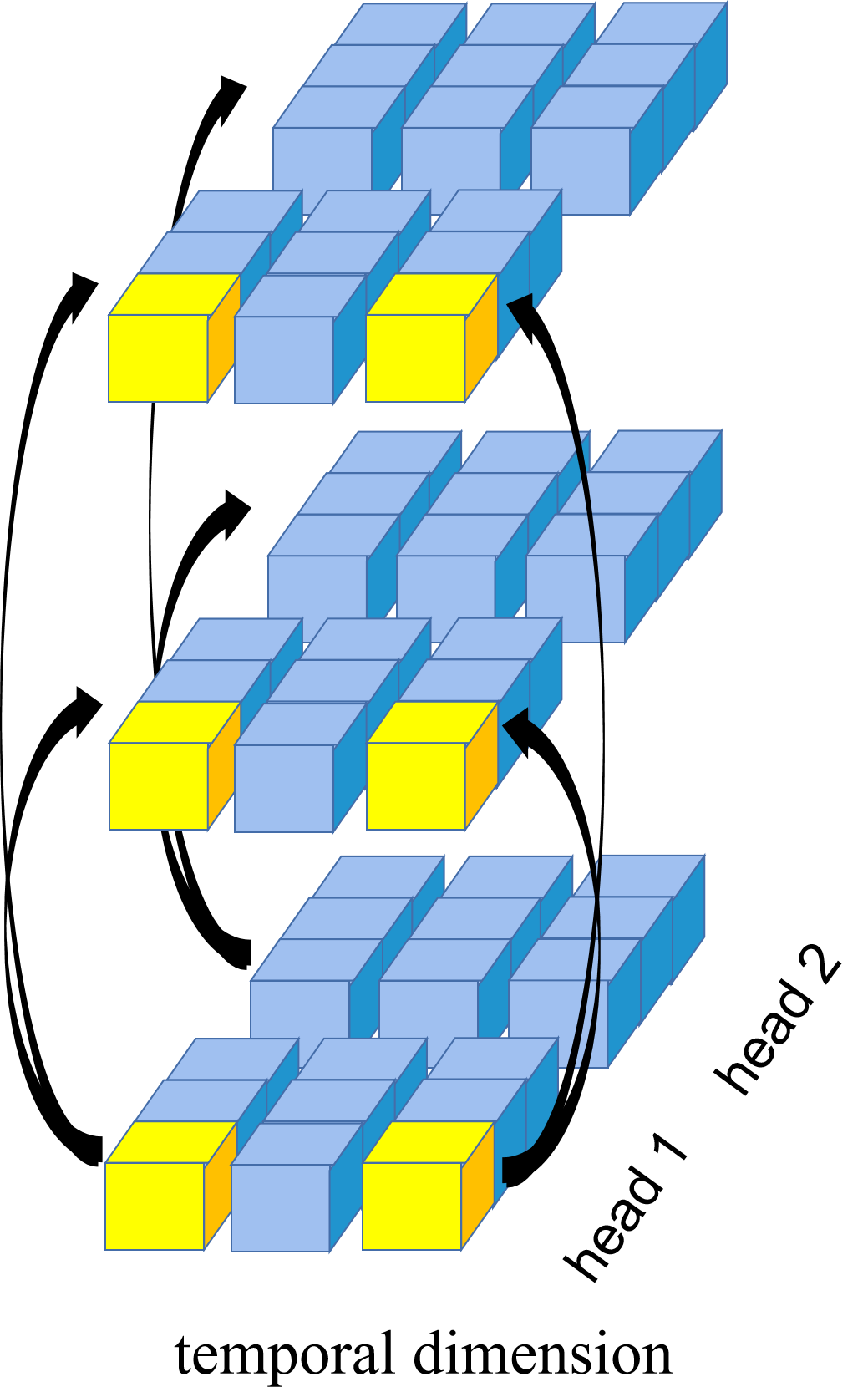}}
%  \vspace{1.5cm}
  \centerline{(b) Cross-lead self-attention.}\medskip
\end{minipage}

\caption{Lead-orthogonal attention mechanism. }
% 这个多写点字，就不居中了
% 看起来不很清晰
\label{fig:attention}

\end{figure}

Based on the above segmentation and pooling procedure, our model is built on both vanilla self-attention blocks and novel lead-orthogonal attention blocks \cite{ding2022davit,dong2022cswin}. The vanilla multihead self-attention (MSA) blocks are cascaded after the lead-orthogonal attention blocks. 
Our lead-orthogonal attention module includes lead-internal multihead self-attention and cross-lead multihead self-attention as shown in \textbf{Fig. \ref{fig:attention}}. It has a clear physical meaning with each group naturally corresponding to segments within the same lead or time period.
In the lead-internal multihead self-attention, we treated all tokens of different leads at the same time, whether they are from initial records or pooled records, as a group. Correspondingly, in the cross-lead multihead self-attention, we treated all tokens in the same lead as a group. 

The group multi-head self-attention in the dual attention block can be represented by
\begin{equation}
    {A_{group}}(Q,K,V) = \{ {A_{Mhead}}({Q_i},{K_i},{V_i})\} _{i = 0}^{{N_{group}}}
\end{equation}
% 这儿插入一些公式。
where attention in a single group is: 
\begin{equation}
    {{A_{Mhead}}\left( {{Q_i},{K_i},{V_i}} \right) = Concat\left\{ {hea{d_j}} \right\}_{group = i}^{j \in [1,{N_{head}}]}}
\end{equation}
and for each attention head: 
\begin{equation}
hea{d_j} = softmax\left( {\frac{{{Q_j}K_j^T}}{{\sqrt {{d_h}} }}} \right){V_j}
    % {hea{d_j} = Attention\left( {{Q_j},{K_j},{V_j}} \right) = softmax\left( {\frac{{{Q_j}K_j^T}}{{\sqrt {{d_h}} }}} \right){V_j}}
\end{equation}

% 公式
% 公式
% 减少复杂度这里不说了。

% Current ECG transformers adopt vanilla transformer block calculating attention on the whole cropped ECG, or alternatively use 1D Swin block, focusing on windows split in the corresponding beat. Our group attention is more similar to window attention in the Swin block but has two orthogonal dimensions and takes the same lead or same time period as a group.

% 这儿可能要插一些公式。和相关的文字。
% 减少计算复杂度这一点，有点弱，要不要提还得另说。

\subsection{Fusion of meta-information}

Previous studies in the field of image classification have shown that the effective use of meta-information can improve the knowledge extraction ability of the model\cite{diao2022metaformer}. 
% As a general method in visual and natural language processing, 
The transformer has the advantage of being able to flexibly adapt to the various input information, thus allowing for intuitive integration of previously discarded meta-information into the proposed model. The meta-information embeddings, combined with the output features of the lead-orthogonal attention layers and the class token, become the input of the following vanilla attention layers.
% Specifically, we linearly projected the meta-information into the feature embeddings as shown in Figure 2. 
% We used positional embeddings to distinguish the embedding of words with different meanings. 

\section{Experiments}

\subsection{Datasets and preprocessing}

PTB-XL is a publicly available clinical dataset containing 21,837 records of 12-lead ECG dataset from 18,885 patients with a 10-second duration \cite{wagner2020ptb}. Each record states whether the recording belongs to one or more multi-label classification tasks including ‘all’, ‘diagnostic’, ‘superdiagnostic’, ‘subdiagnostic’, ‘rhythm’ and ‘form’. Meta-information including age, gender, height and weight is available. 

Chapman is another open-source clinical database comprised of 12-lead ECG signals from 10646 individuals. Age and gender are available in this dataset \cite{zheng202012}. The labels of all data are clustered into four disjoint categories SR, SB, GSVT, and AFIB. Both the original and denoised versions are provided. We trained on both original and denoised records and tested only on original records.

We selected the 100Hz version for both datasets. For PTB-XL, according to the recommended partition, eight of the 10 folds are used for training, one for validation and the other for testing. For Chapman, we implemented 10 fold cross-validation with 80\% indices for training, 10\% for validation and 10\% for testing.

\subsection{Evaluation metrics}

\begin{table}\centering
\caption{Ablation study results(\%). }
  \begin{tabular}{ccccc}
  
    \toprule[1.5pt]
    %%%%%%%%%%%%%%%%%%%%%%%%%%%%%%%%%%%%%%%%%%%%%
    %\textwidth 是每一行的宽度.[0.1\textwidth]设定单元格宽度
    % [c]  单元格文本居中对齐
    % {name} 单元格内容
    %%%%%%%%%%%%%%%%%%%%%%%%%%%%%%%%%%%%%%%%%%%%%
    \makebox[0.1\textwidth][c]{\makecell[c]{Lead \\ independence}} &
    \makebox[0.04\textwidth][c]{\makecell[c]{Dual \\ scale}} & 
    \makebox[0.06\textwidth][c]{\makecell[c]{Multiple \\ attention}} & 
    \makebox[0.08\textwidth][c]{\makecell[c]{Meta- \\ information}} & 
    \makebox[0.06\textwidth][c]{\makecell[c]{macro \\ AUC}}        
    \\
    \midrule[1pt]
    
    \ding{53}&

    \ding{53}&

    \ding{53}&

    \ding{53}&

    91.6

    \\

    \ding{51}&

    \ding{53}&

    \ding{53}&

    \ding{53}&

    92.6

    \\
    \ding{51}&

    \ding{51}&

    \ding{53}&

    \ding{53}&

    92.9

    \\

    \ding{51}&

    \ding{53}&

    \ding{51}&

    \ding{53}&

    92.8

    \\
    \ding{51}&

    \ding{51}&

    \ding{51}&

    \ding{53}&

    93.4

    \\
    \ding{51}&

    \ding{51}&

    \ding{51}&

    \ding{51}&

    \textbf{94.2}

    \\

    \bottomrule[1.5pt]
  \end{tabular}
  \label{table: ablation exp}
\end{table}

The comparison with current methods on PTB-XL dataset was achieved through the macro-AUC, which measures class separability by calculating the area under the receiver operating characteristic curve for each label and then averaging them. 
% A higher AUC indicates better class separation ability. 
For the Chapman database, we further adopted the following metrics: accuracy, macro-precision, macro-recall and macro-F1. 
% We calculate these metrics for all the classes and The accuracy is the proportion of correctly predicted samples in the total samples. 
For each class we have 
\begin{equation}
    {Precision = \frac{{TP}}{{TP + FP}}}
\end{equation}
\begin{equation}
    {Recall = \frac{{TP}}{{TP + FN}}}
\end{equation}
\begin{equation}
    {F1 = \frac{{2 * Recall * Precision}}{{Recall * Precision}}}
\end{equation}
The macro metrics calculates the average of results for each class.
% 公式
% 公式
% 公式
% The macro metrics calculates the average of results for each class.

\subsection{Implementation details}

\begin{table*}[t]\centering
\caption{Experimental results(\%) on PTB-XL. }
  \begin{tabular}{ccccccc}
  
    \toprule[1.5pt]
    %%%%%%%%%%%%%%%%%%%%%%%%%%%%%%%%%%%%%%%%%%%%%
    %\textwidth 是每一行的宽度.[0.1\textwidth]设定单元格宽度
    % [c]  单元格文本居中对齐
    % {name} 单元格内容
    %%%%%%%%%%%%%%%%%%%%%%%%%%%%%%%%%%%%%%%%%%%%%
    \makebox[0.2\textwidth][c]{method} &
    \makebox[0.1\textwidth][c]{all} & 
    \makebox[0.1\textwidth][c]{diag} & 
    \makebox[0.1\textwidth][c]{superdiag} & 
    \makebox[0.1\textwidth][c]{subdiag} & 
    \makebox[0.1\textwidth][c]{rhythm} & 
    \makebox[0.1\textwidth][c]{form}             
    \\
    \midrule[1pt]
    Wavelet+NN \cite{strodthoff2020deep}&
     % \cite{strodthoff2020deep}&
 
    84.9&
 
    85.5&
 
    87.4&
 
    85.9&
 
    89.0&
 
    75.7
    \\
 
    LSTM \cite{strodthoff2020deep}&
     % \cite{strodthoff2020deep}&
 
    90.7&
 
    92.7&
 
    92.7&
 
    92.8&
 
    95.3&
 
    85.1
    \\
 
    LSTM\_bidir \cite{strodthoff2020deep}&
     % \cite{strodthoff2020deep}&
 
    91.4&
 
    93.2&
 
    92.1&
 
    92.3&
 
    94.9&
 
    87.6
    \\
 
    CNN+LSTM \cite{mehari2022self}&
     % \cite{mehari2022self}&
 
    93.2&
 
    93.2&
 
    92.9&
 
    92.9&
 
    95.0&
 
    89.2
    \\
 
    resnet1d \cite{strodthoff2020deep}&
     % \cite{strodthoff2020deep}&
 
    91.9&
 
    93.6&
 
    93.0&
 
    92.8&
 
    94.6&
 
    88.0
    \\
 
    method in \cite{wang2021automated}&
     % \cite{wang2021automated}&
 
    -&
 
    -&
 
    93.1&
 
    -&
 
    -&
 
    -
    \\
 
    ST-CNN-GAP-5 \cite{anand2022explainable}&
     % \cite{anand2022explainable}&
 
    -&
 
    -&
 
    \textbf{93.4}&
 
    -&
 
    -&
 
    -
    \\
 
    fcn \cite{strodthoff2020deep}&
     % \cite{strodthoff2020deep}&
 
    91.8&
 
    92.6&
 
    92.5&
 
    92.7&
 
    93.1&
 
    86.9
    \\
 
    IMLE-Net \cite{reddy2021imle}&
     % \cite{reddy2021imle}&
 
    -&
 
    -&
 
    92.2&
 
    -&
 
    -&
 
    -
    \\
 
    inception1d \cite{strodthoff2020deep}&
     % \cite{strodthoff2020deep}&
 
    92.5&
 
    93.1&
 
    92.1&
 
    \underline{93.0}&
 
    95.3&
 
    \underline{89.9}
    \\
 
    xresnet1d \cite{strodthoff2020deep}&
     % \cite{strodthoff2020deep}&
 
    92.5&
 
    \underline{93.7}&
 
    92.8&
 
    92.9&
 
    95.7&
 
    89.6
    \\
 
    multi-period attention \cite{zhang2021multi}&
     % \cite{zhang2021multi}&
 
    92.7&
 
    93.1&
 
    93.0&
 
    92.3&
 
    \textbf{97.2}&
 
    85.2
    \\
 
    ViT \cite{li2021bat}&
     % \cite{li2021bat}&
 
    86.2&
 
    -&
 
    -&
 
    -&
 
    -&
 
    -
    \\
 
    % Swin-ViT\cite{li2021bat}&
    %  % \cite{li2021bat}&
 
    % 89.6&
 
    % -&
 
    % -&
 
    % -&
 
    % -&
 
    % -
    % \\
 
    BaT \cite{li2021bat}&
     % \cite{li2021bat}&
 
    90.5&
 
    -&
 
    -&
 
    -&
 
    -&
 
    -
    \\
    % \makebox[0.1\textwidth][c]{\makecell[c]{DLTM-ECG \\ (meta removed)}} &
    % DLTM-ECG(meta removed)&
 
    % \underline{93.4}&
 
    % 93.5&
 
    % 93.0&
 
    % \underline{93.4}&
 
    % 95.8&
 
    % 87.8
    % \\
 
    \textbf{DLTM-ECG}&
 
    \textbf{94.2}&
 
    \textbf{93.9}&
 
    \underline{93.2}&
 
    \textbf{93.8}&
 
    \underline{96.1}&
 
    \textbf{91.7}
    \\

    \bottomrule[1.5pt]
  \end{tabular}
  \label{table: ptb-xl exp}
\end{table*}

% \begin{table*}[t]\centering
% \caption{Experimental results(\%) on Chapman. }
%   \begin{tabular}{cccccc}
  
%     \toprule[1.5pt]
%     %%%%%%%%%%%%%%%%%%%%%%%%%%%%%%%%%%%%%%%%%%%%%
%     %\textwidth 是每一行的宽度.[0.1\textwidth]设定单元格宽度
%     % [c]  单元格文本居中对齐
%     % {name} 单元格内容
%     %%%%%%%%%%%%%%%%%%%%%%%%%%%%%%%%%%%%%%%%%%%%%
%     \makebox[0.2\textwidth][c]{method} &
%     \makebox[0.1\textwidth][c]{accuracy} & 
%     \makebox[0.1\textwidth][c]{macro precision} & 
%     \makebox[0.1\textwidth][c]{macro recall} & 
%     \makebox[0.1\textwidth][c]{macro F1} & 
%     \makebox[0.1\textwidth][c]{macro AUC}        
%     \\
%     \midrule[1pt]
    
%     method in\cite{yildirim2020accurate}&

%     96.1&

%     95.8&

%     95.4&

%     95.6&

%     -

%     \\

%     ST-CNN-GAP-5\cite{anand2022explainable}&

%     96.2&

%     95.9&

%     95.7&

%     95.8&

%     99.5

%     \\
    
%     \textbf{DLTM-ECG}&
 
%     \textbf{96.3}&
 
%     \textbf{96.2}&
 
%     \textbf{95.7}&
 
%     \textbf{95.9}&
 
%     \textbf{99.7}

%     \\

%     \bottomrule[1.5pt]
%   \end{tabular}
%   \label{table: chapman exp}
% \end{table*}

\begin{table}[t]\centering
\caption{Experimental results(\%) on Chapman. }
  \begin{tabular}{cccc}
  
    \toprule[1.5pt]
    %%%%%%%%%%%%%%%%%%%%%%%%%%%%%%%%%%%%%%%%%%%%%
    %\textwidth 是每一行的宽度.[0.1\textwidth]设定单元格宽度
    % [c]  单元格文本居中对齐
    % {name} 单元格内容
    %%%%%%%%%%%%%%%%%%%%%%%%%%%%%%%%%%%%%%%%%%%%%
    \makebox[0.1\textwidth][c]{metric} &
    \makebox[0.07\textwidth][c]{\cite{yildirim2020accurate}} & 
    \makebox[0.07\textwidth][c]{\cite{anand2022explainable}} & 
    \makebox[0.1\textwidth][c]{\textbf{DLTM-ECG}}       
    \\
    \midrule[1pt]
    
    accuracy&

    96.1&

    96.2&

    \textbf{96.3}

    \\

    macro precision&

    95.8&

    95.9&

    \textbf{96.2}

    \\
    
    macro recall&
 
    95.4&
 
    95.7&
 
    \textbf{95.7}

    \\
    
    macro F1&
 
    95.6&
 
    95.8&
 
    \textbf{95.9}

    \\
    
    macro AUC&
 
    -&
 
    99.5&
 
    \textbf{99.7}

    \\

    \bottomrule[1.5pt]
  \end{tabular}
  \label{table: chapman exp}
\end{table}

In each training step, a sliding window with a random offset generates segments from the current batch data, and DLTM-ECG iterates over the segments in turn.
This avoids unstable training caused by excessive randomness and alleviates inefficiencies caused by the transformer's lack of inductive bias to spatial invariance. 
% In each training step, DLTM-ECG iterates over the segments intercepted from the current batch data in turn.
We warmed up the model for 20 epochs at a linear schedule and then applied ReduceLROnPlateau learning rate schedule, which automatically halves the learning rate when the validation score doesn't improve. The max learning rate was set to $0.003$, and the batch size is set to 128 for both datasets. An Adamw optimizer with ${\beta _1=0.9}$, ${\beta _2=0.99}$ and weight decay $ = 0.05$ was applied as our base optimizer. Then we wrapped the base optimizer with the SAM method, and the radius of hypersphere ${\rho}$ was set to 0.05 \cite{foret2020sharpness}. An exponential moving average (EMA)\cite{he2022masked} with ${\alpha=0.998}$ is implemented, and the shadow weight is updated after each training step. BCEWithLogits loss function is used.%\cite{paszke2019pytorch}.

% and the batch size was set to 128 for both datasets. An Adam optimizer with ,  was applied as our base optimizer, and its weight decay was set to 0.05. An exponential moving average (EMA) [23] with  for PTB-XL or  for Chapman was implemented, and the shadow weight was updated after each training step. Dropout was set to 0.2, and BCEWithLogits loss function was used [33]. Some model structure hyperparameters were set as shown in Table 1. 

% Specifically, for an input array of in each batch with batch size , we intercept seven segments of the array in turn, and the number of points in each segment is . The starting point of the segment i is sampled randomly in the following closed intervals:

% 这儿看着说一下。
% 不要太那啥

\subsection{Ablation study}

We provided ablation experiments on the most fine-grained task of PTB-XL. Except for ablation variables, other training settings remained the same. 
For meta-information ablation, we simply removed these tokens. 
For multi-attention ablation, we replaced lead-orthogonal attention blocks with double numbers of multihead self-attention blocks that have a similar number of parameters. 
% The class token is initialized randomly and concatenated to the first layer of the model. 
To remove the dual scale, we skipped the pooling step. 
We ablated the independence of different leads in this way: instead of dividing 50 points (i.e. 0.5 seconds) off one lead into one patch, we took the 12-channel record of 0.05 seconds as the basic unit, which divides the record temporally and keeps a similar number of patch embeddings. 
Ablation results shown in \textbf{Table \ref{table: ablation exp}} demonstrate the effectiveness of each innovation we propose.

\subsection{Comparison with the state-of-the-arts}

The evaluation results on PTB-XL are shown in \textbf{Table \ref{table: ptb-xl exp}} together with current SOTA methods. 
% try some extra words if they add to anothor page
A previous approach provides benchmarking results covering models including LSTM\_bidir, inception1d and xresnet1d \cite{strodthoff2020deep}. 
Periodic masks were introduced in \cite{zhang2021multi} to better capture rhythm information. 
Non-local convolutional block attention module was used in [17] to capture long-range correlation. 
Both spatial and temporal layers were adopted in \cite{anand2022explainable} which lead to a diagnostic model with decent performance. 
% Methods designed in [8] and [17] also give advanced classification score on diagnostic task. 
The work in \cite{mehari2022self} combines CNN and LSTM and generalizes well on multiple tasks. 
% None of these methods use meta-information, which can be easily obtained from the databases.
% None of these methods use meta-information, which can be easily obtained from the databases.
We run our experiment 10 times and report the average macro-AUC for each task. Our approach achieved the most advanced overall performance on multiple tasks and exceeded the baseline ViT \cite{li2021bat} by 8\% with an AUC score of 94.2\%. Even without the addition of meta-information, the model still obtains well performance, which proves the effectiveness of our method for feature extraction.
We also list the results of the SOTA end-to-end deep learning methods using Chapman in \textbf{Table \ref{table: chapman exp}} together with our experimental results under the same settings as PTB-XL. Our method shows good generalization performance in multiple metrics and obtains scores matching or performing better than compared models.

\section{Conclusion}

In this work we present a transformer-based model DLTM-ECG to classify clinical ECG records. We considered characteristics of the transformer and ECG signals to design specific structures. Lead-separated segmentation was proposed to capture the correlation between leads, and the signal was pooled to obtain dual-scale representation. Multiple attention mechanisms further improved the feature extraction ability. The incorporation of meta-information into the model is an important innovation that not only improves classification scores but also reveals the potential to utilize multimodal data in a variety of different physiological signal analyses. Classification results on multiple tasks and ablation experiments have confirmed the effectiveness of the proposed technology.

\vfill\pagebreak

% \section{REFERENCES}
% \label{sec:refs}

% List and number all bibliographical references at the end of the
% paper. The references can be numbered in alphabetic order or in
% order of appearance in the document. When referring to them in
% the text, type the corresponding reference number in square
% brackets as shown at the end of this sentence \cite{C2}. An
% additional final page (the fifth page, in most cases) is
% allowed, but must contain only references to the prior
% literature.


\bibliographystyle{IEEEbib}
\bibliography{refs.bib}

\end{document}